\documentclass{article}

\usepackage{amsmath,amssymb}
\usepackage{hyperref}
\usepackage{graphicx}
\usepackage{algorithm}
\usepackage{algpseudocode}
\usepackage{booktabs}
\usepackage{multirow}
\usepackage[table]{xcolor}
\usepackage{xcolor} 
\usepackage[final]{corl_2026} 
\usepackage{float}

\usepackage{cleveref}
\crefname{appendix}{Appendix}{Appendices}
\Crefname{appendix}{Appendix}{Appendices}

\usepackage{wrapfig}
\usepackage{titletoc}
\usepackage{subcaption}

\title{Start Right, Arrive Right: Asynchronous Execution via Initial Noise Selection}

\author{
  \parbox{\textwidth}{\centering
  \vspace{0.2in}
    Trong-Bao Ho$^{1,2\ast}$ \quad
    Quang-Tan Nguyen$^{1\ast}$ \quad
    Thien-Loc Ha$^{1\ast}$ \quad
    Gia-Binh Nguyen$^{1,2}$ \quad \\[0.05in]
    Viet-Thanh Nguyen$^{1}$ \quad
    Long Dinh$^{1,2}$ \quad
    Minh N. Vu$^{1,2}$ \quad
    Duy M. H. Nguyen$^{3,4,5}$ \\[0.05in]
    An T. Le$^{1,2,6}$ \quad
    Ngo Anh Vien$^{1,2}$ \\[1.2em]
    \footnotesize\normalfont
    $^{1}$VinRobotics \quad
    $^{2}$Center for AI Research \& CECS, VinUniversity \quad
    $^{3}$DFKI \quad
    $^{4}$University of Stuttgart \\[0.05in]
    $^{5}$Max Planck Research School for Intelligent Systems (IMPRS-IS)\quad
    $^{6}$IAS, TU Darmstadt \\[0.05in]
    \footnotesize\normalfont $^{\ast}$Equal Contributors.
  }
}

\usepackage{xcolor}

\definecolor{gray}{cmyk}{0, 0, 0, 0.1}

\begin{document}
\maketitle


\begin{abstract}
    Action chunking enables robot policies to produce temporally coherent behavior, but generating multi-step action sequences with flow-based policies incurs latency that is incompatible with real-time control. Under asynchronous execution, the robot continues executing the current chunk while the next one is generated, causing even minor delays to create inconsistencies at chunk boundaries. Existing methods address this problem by steering generation toward the already executed action prefix.  We instead show that prefix consistency can be achieved by selecting an appropriate initial noise before generation begins, allowing the unmodified flow ODE to produce a coherent next chunk. This establishes initial-noise selection as a complementary approach to gradient-based asynchronous execution. We introduce \textbf{PAINT}, a training-free method that finds this noise via backward Euler inversion and constructs the final chunk through a repainting rule. In summary, \texttt{PAINT} requires no gradients, retraining, or policy modification; and achieves task performance comparable to RTC while improving prefix consistency across \textit{12 simulated benchmarks} and \textit{6 real-world manipulation tasks} spanning single-arm, bimanual, and humanoid embodiments.
    Project site is available at: ~\href{https://paint-action-chunking.github.io}{\texttt{https://paint-action-chunking.github.io}}.
\end{abstract}



\section{Introduction}
\label{sec:intro}

\begin{wrapfigure}[14]{R}{0.48\linewidth}
    \centering
    \vspace{-4pt}
    \includegraphics[width=.92\linewidth]{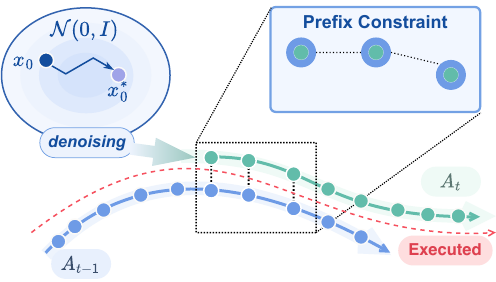}
    \vspace{-4pt}
    \caption{The \emph{prefix constraint}: the first $d$ actions of chunk $A_t$ must
    approximate the last $d$ actions of chunk $A_{t-1}$. A carefully chosen initial noise
    satisfies this constraint without modifying the policy.}
    \label{fig:intro}
\end{wrapfigure}
Flow matching~\citep{lipmanflow} and diffusion~\citep{ho2020denoising} policies have
achieved remarkable dexterity by predicting \emph{action chunks}, sequences of future
actions generated in a single forward pass~\citep{zhao2023learning, chi2025diffusion,
black2024pi_0}. Action chunking improves temporal coherence, but each chunk requires
multiple sequential denoising steps to generate. Under asynchronous inference, the robot
cannot wait: it continues executing the previous chunk, and by the time the new one is
ready, it has already advanced $d$ steps. This creates the \emph{prefix constraint}:
those $d$ actions must be consistent with what was just executed, or the robot
experiences a discontinuous jump at the chunk boundary. Without enforcement, performance
degrades substantially with delay, and this gap that widens as larger models push inference
latency beyond the controller's sampling period.

Enforcing the prefix constraint is therefore critical, and the natural response is to
steer generation toward the prefix target \emph{during} denoising. Prior work does
exactly this, using backpropagation through the policy~\citep{black2025real}, model
retraining~\citep{black2025training, tang2025vlash}, or additional sampling
compute~\citep{liu2024bidirectional} to enforce continuity at generation time. Yet, this
framing raises a deeper question: \textit{``What if the trajectory never needed to be corrected in
the first place?"} \textit{``What if the right initial noise, chosen \emph{before} generation begins,
would naturally produce a consistent prefix?"}

Fortunately, flow matching offers a cleaner path. Optimal-transport flow
matching~\citep{lipmanflow} (OT-FM) learns a transport map with an approximately
\emph{local} structure: each output position is governed largely by the noise at the
corresponding input position, a property observed empirically in diffusion models
by~\citet{mao2023guided} and supported theoretically for OT-FM by~\citet{tongimproving}. This locality has a direct consequence: if the prefix of a generated chunk depends mainly on the prefix of the initial noise, then inverting the ODE from a desired prefix target recovers a noise that satisfies the prefix
constraint under the unmodified forward pass, without velocity correction or gradients
required. This reframes the prefix constraint as a \emph{noise-selection problem} rather
than a trajectory-steering problem.


This insight leads directly to our method named \textbf{\texttt{\texttt{PAINT}}} (\textbf{P}refix-\textbf{A}nchored
\textbf{IN}i\textbf{T}ial noise), a training-free method that enforces the prefix
constraint by inverting the flow ODE to find an initial noise $x_0^\ast$, then
constructs the full chunk via the re-painting principle of~\citet{mao2023guided}.
Prefix and suffix of an action chunk are generated jointly from this re-painted noise, producing a continuation that is approximately consistent with the policy distribution under the executed prefix. Unlike velocity-steering approaches, PAINT enforces this consistency through initial-noise selection before the final denoising trajectory is generated.
\texttt{\texttt{PAINT}} requires (i) \textit{no modification to the base policy, (ii) no training data, and (iii) no
auto-differentiation framework at deployment}. This makes it compatible with forward-only inference runtimes and graph-compilation pipelines.

Our contributions are: (i)~introducing initial-noise selection as a complementary mechanism for asynchronous execution, grounded in the locality structure of OT flow matching; (ii)~\texttt{\texttt{PAINT}}, a
training-free, backpropagation-free inference-time method that matches Real-Time Chunking (RTC)~\cite{black2025real} in task success while consistently improving prefix consistency; and (iii)~an empirical
demonstration that \texttt{\texttt{PAINT}} transfers to any pretrained flow matching policy without
retraining or modification, validated across $12$ simulated benchmarks and six
real-world tasks on two VLA architectures and three robot embodiments.
\section{Related Work}
\label{sec:related_work}
\vspace{-0.1in}
\texttt{PAINT} sits at the intersection of two active research areas: real-time execution of
action-chunking policies and structured manipulation of initial noise in generative
models. We review each in turn.

\textbf{Real-Time Execution of Action-Chunking Policies.}
Action chunking -- predicting a fixed-length sequence of future actions in an
inference call -- has become the dominant paradigm for robot manipulation~\citep{zhao2023learning, chi2025diffusion}, and large-scale
VLAs~\citep{black2024pi_0, bjorck2025gr00t,
intelligence2025pi05visionlanguageactionmodelopenworld, kim2024openvla,
shukor2025smolvla} scale these architectures at the cost of inference latency.
\citet{black2025real} provide a detailed treatment of this latency problem; their
RTC framework is our primary baseline. Existing methods address the gap between inference and execution at different stages. TT-RTC~\citep{black2025training} and VLASH~\citep{tang2025vlash} fine-tune the policy to tolerate delay, while Streaming Diffusion and Streaming Flow Policies~\citep{hoeg2024streaming,jiang2025streaming} introduce training schemes for faster inference.
At inference time, BID~\citep{liu2024bidirectional} uses candidate filtering through rejection sampling, while A2C2~\citep{sendai2025leave}, SAIL~\citep{arachchige2025sail}, FASTER~\citep{lu2026faster}, REMAC~\citep{wang2026real} and SoftRTC~\citep{liu2026action} modify the generated action trajectory through correction, guidance, sampling
modification, or relaxed prefix-consistency objectives. ABPolicy~\citep{yang2026abpolicy} enforces continuity with a B-spline action representation, and DiscreteRTC~\citep{wang2026discretertc} uses discrete diffusion to align iterative generation with asynchronous execution better. Despite their diversity, these methods generally intervene after the
initial noise is sampled, either at training time, during denoising, or on the generated output, but never on the noise itself. \texttt{PAINT} acts at exactly that earlier point. A
concurrent survey~\citep{agouzoul2026understanding} compares several of these baselines
but omits the noise-space perspective we adopt.

\textbf{Initial Noise Manipulation and Inversion in Generative Models.}
The initial noise vector, traditionally treated as unstructured randomness, in fact
carries semantic structure. \citet{mao2023guided, mao2024lottery} demonstrated
spatially localized generation in vision: perturbing the noise at position $i$ primarily
affects the output at $i$. \citet{patil2026you} extended this to robot policies, showing
that a carefully chosen constant noise vector improves frozen-policy performance. Noise
manipulation has also been approached through learning: \citet{wagenmakersteering} train
a noise-space policy via RL, RTI-DP~\citep{duan2025rti} warm-starts denoising from the
previous prediction, A2A~\citep{jia2026a2a} replaces noise with proprioceptive
embeddings, and UniSteer~\citep{lu2026unisteer} inverts a flow decoder for RL-based
adaptation.

Inversion methods recover the initial noise for a desired output: DDIM
inversion~\citep{songdenoising} runs the generative process backward, with later work
improving its accuracy~\citep{meiri2023fixed, pan2023effective, zhang2024exact,
hong2024exact}, and flow matching admits single-step inversion~\citep{li2026reverse}.
\texttt{PAINT} is the first to apply ODE inversion to the prefix constraint in asynchronous
action-chunk inference, requiring no learning, retraining, or human feedback, and then
combining backward Euler inversion~\citep{pan2023effective} with the re-painting principle
of~\citet{mao2023guided} to produce temporally coherent chunks.
\section{Preliminaries and Motivation}
\label{sec:background}

We adopt the notation and problem formulation of~\citet{black2025real}. Given an observation $o_t$ and initial noise $x_0$, a deterministic policy $\pi_\theta(o_t, x_0)$ outputs an \emph{action chunk} $A_t^{0:H-1}$ of \emph{action horizon} $H$. Rather than waiting for the chunk to be fully consumed, the robot executes only the first $s$ actions $A_t^{0:s-1}$ before issuing the next inference call; we refer to $s$ as the \emph{execution horizon}. Since each inference call requires wall-clock time $\delta$ to complete $N$ denoising steps, and the controller operates at period $\Delta t$, the robot advances $d = \lfloor \delta / \Delta t \rfloor$ timesteps during chunk generation; we call $d$ the \emph{inference delay}. Real-time execution requires $d \leq H - s$, guaranteeing that a new chunk is ready before the current one is exhausted. Consequently, by the time chunk $A_t$ becomes available, the robot has already consumed $s + d$ actions from the previous chunk $A_{t-1}^{0:s+d-1}$.

\textbf{The Problem of Prefix Constraints in Asynchronous Inference.}
In synchronous inference, the policy $\pi_\theta$ computes the next chunk $A_t$ only
after finishing execution of $A_{t-1}$. In asynchronous inference, by contrast, the robot executes
$A_{t-1}$ while $A_t$ is being computed, so the next chunk must satisfy the following \emph{prefix constraint} to ensure motion continuity:
\begin{equation} \label{eq:prefix_constraint}
    A_{t-1}[s+i] = A_{t}[i] \quad \text{for}\ i = 0, 1,\dots, d-1 \,.
\end{equation}
Violating it causes a discontinuous jump at the chunk
boundary, producing jerky or unsafe motion.


\textbf{Initial Noise as a Control Variable.}
When the velocity field $v_\pi(\cdot,o,\tau)$ is Lipschitz continuous in its first argument and continuous in $\tau$, the continuous-time flow ODE defines a unique flow map from $x_0$ to $x_1$~\citep{lipmanflow, lipman2024flow}. Under these idealized conditions, the map is invertible. In practice, however, learned policies are evaluated with finite-step numerical solvers and the target prefix may not lie exactly on the learned trajectory manifold. We therefore treat \texttt{PAINT} as an approximate inverse procedure: it only needs to recover an initial noise whose forward rollout produces a prefix sufficiently close to the executed actions.
This makes $x_0$ a natural control variable: some output constraints can be addressed by searching over the initial noise $x_0$ appropriately, without modifying the velocity field $v_\pi(x_\tau, o, \tau)$ at any denoising step~\citep{mao2024lottery}. Recent work corroborates this view:~\citet{wagenmakersteering} and~\citet{patil2026you} demonstrate that structured noise choices steer frozen policies toward desired behaviors, and~\citet{jiang2025streaming} bias the initial noise toward a running estimate to encourage smooth inter-chunk transitions. This motivates our approach: the prefix constraint (\Cref{eq:prefix_constraint}) defines a set in noise space, and satisfying it approximately can be approached by searching for an appropriate point $x_0^\ast$ within that set.
\section{Asynchronous Inference as a Noise Selection Problem}
\label{sec:proposed_method}

\begin{figure}[t]
    \centering
    \includegraphics[width=\linewidth]{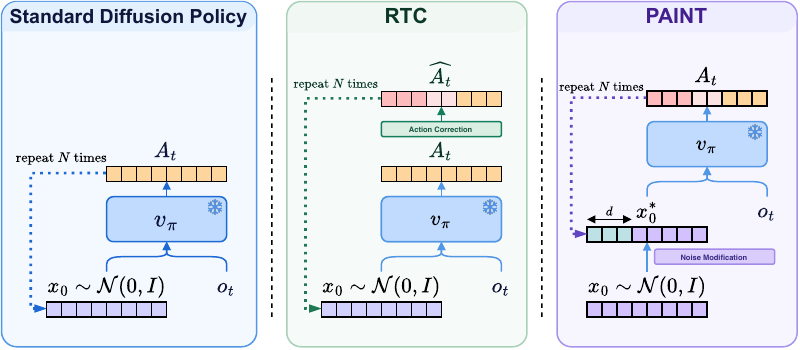}
    \caption{Overview of standard diffusion policy~\citep{chi2025diffusion, black2024pi_0, bjorck2025gr00t} (a), RTC~\citep{black2025real} (b), and our proposed method PAINT (c), which leverages modified noise. Given a frozen, pretrained diffusion or flow-matching policy $\pi_\theta$, instead of modifying the ODE at each denoising step, we modify the initial noise $x_0$ to get $x_0^\ast$, which satisfies the prefix constraint (\Cref{eq:prefix_constraint}). We show that $x_0^\ast$ can improve policy performance while satisfying the prefix constraint across model architectures and embodiments.}
    \label{fig:overview}
\end{figure}

Existing methods such as RTC~\citep{black2025real} and BID~\citep{liu2024bidirectional} enforce the prefix constraint by steering the velocity field $v_\pi$ at every denoising step (see~\Cref{fig:overview}b), accepting backpropagation or heavy compute as the price. We take a different route: under OT-FM, there exists an initial noise $x_0^\ast$ such that the \emph{unmodified} ODE already satisfies the prefix constraint. The problem reduces to finding $x_0^\ast$. Once found, inference runs exactly as during training: no guidance, no backpropagation, no velocity correction. We call this method \textbf{PAINT} (\textbf{P}refix-\textbf{A}nchored \textbf{IN}i\textbf{T}ial noise) (see~\Cref{fig:overview}c). The following subsections derive \texttt{PAINT}: we first explain why it should be noise selection, then show how ODE inversion can find $x_0^\ast$ efficiently.
\subsection{From Velocity Steering to Noise Selection}

Asynchronous inference can be addressed in two directions: modifying the ODE \emph{during} generation to steer toward the prefix, or selecting the initial noise \emph{before} generation so that the ODE automatically satisfies the constraint. Existing inference-time methods, including RTC~\citep{black2025real}, operate in the first direction. We argue that the second direction is both simpler and less biased toward the learned distribution. Simpler, because it requires no change to the velocity field $v_\pi$ at any ODE step, the standard solver runs unaltered, with complexity absorbed entirely into a selection of the initial noise $x_0$. Less biased, because the trajectory is generated without modifying the learned flow field: since $v_\pi$ is never corrected, the generated chunk more closely matches samples induced by the learned flow rather than the endpoint of a steered path that may lie outside the support of the learned distribution.

\textbf{Velocity Steering May Deviate from Learned Flow.}
RTC applies a pseudoinverse correction to the velocity field at each ODE step:
\begin{equation} \label{eq:rtc}
    v_\pi^{\text{RTC}}(A_t^\tau, o_t, \tau) 
    \;=\; v_\pi(A_t^\tau, o_t, \tau) 
    \;+\; w \cdot J^\dagger \cdot \bigl(Y - \widehat{A_t^1}\bigr) \,.
\end{equation}
where $J = \partial \widehat{A_t^1} / \partial A_t^\tau$ is the Jacobian of the denoiser prediction and $J^\dagger$ is its pseudoinverse. The correction pushes the predicted output $\hat{x}_1$ toward the prefix target. However, we decompose the correction into components parallel and orthogonal to the natural velocity field.
\begin{equation} \label{eq:rtc_decomp}
    J^\dagger \cdot e \;=\; \delta v_\parallel \;+\; \delta v_\perp\,, 
    \qquad e = Y - \widehat{A_t^1} \,.
\end{equation} 
the orthogonal component $\delta v_\perp \neq 0$ in general, steering $x_t$ away from the natural ODE trajectory. The correction can move the trajectory away from the dynamics induced by the learned vector field, potentially producing chunks that differ from those generated by the unmodified policy.

\textbf{Noise Selection Leaves the Learned Flow Field Unchanged.}
There exists an initial noise $x_0^\ast \in \mathbb{R}^{H \times D_\text{action}}$ such that the standard unmodified ODE already satisfies the prefix constraint. Running the ODE from $x_0^\ast$ requires no correction to $v_\pi$ at any step. Since \texttt{PAINT} changes only the initial noise and leaves the learned flow field unchanged, the generated chunk is obtained from the same forward dynamics as the base policy, conditioned on the observation $o_t$. The prefix constraint is satisfied not by distorting the distribution but by choosing the right starting point within it. The remainder of this section derives an efficient procedure for finding $x_0^\ast$ as shown in~\Cref{alg:paint} below.

\subsection{PAINT: Prefix-Anchored Initial Noise}
\begin{algorithm}[t]
\caption{\textbf{PAINT} (\textbf{P}refix-\textbf{A}nchored \textbf{IN}i\textbf{T}ial noise)}
\label{alg:paint}
\begin{algorithmic}[1]
\Require Observation $o_t$, executed prefix $A_{t-1}^{s{:}s{+}d{-}1}$, delay $d$, execution $s$, ODE steps $N$
\Ensure Action chunk $A_t$

\State $x_0^{\text{free}} \sim \mathcal{N}(0, I)$
\State $x_1^{\text{naive}} \leftarrow \pi_\theta(x_0^{\text{free}},\; o_t)$
    \Comment{naive forward pass \quad [$N$ calls]}

\State $x_1^{\text{target}} \leftarrow 
    \bigl[A_{t-1}^{s{:}s{+}d{-}1},\; x_1^{\text{naive}}[d{:}]\bigr]$
    \Comment{construct inversion target}

\State $x_\tau \leftarrow x_1^{\text{target}}$
\For{$\tau = 1,\; 1{-}{\frac{1}{N}},\; \ldots,\; {\frac{1}{N}}$}
    \State $x_\tau \leftarrow x_\tau - {\frac{1}{N}} \cdot 
        v_\pi(x_\tau,\; o_t,\; \tau)$
    \Comment{backward Euler \quad [$N$ calls]}
\EndFor
\State $x_0^\text{inv} \leftarrow x_\tau$

\State $x_0^\ast \leftarrow 
    \bigl[x_0^\text{inv}[{:}d],\; x_0^{\text{free}}[d{:}]\bigr]$
    \Comment{Mao re-painting rule}

\State $A_t \leftarrow \pi_\theta(x_0^\ast,\; o_t)$
    \Comment{final forward pass \quad [$N$ calls]}

\State \Return $A_{t}$
\end{algorithmic}   
\end{algorithm}

To find $x_0^\ast$, we run the flow ODE in reverse. Starting from $x_1^{\text{target}}$ at $\tau{=}1$ and applying backward Euler for $N$ steps,
\begin{equation} \label{eq:euler_invert}
x_{\tau-\Delta\tau}
=
x_\tau - \Delta\tau \, v_\pi(x_\tau,o,\tau),
\qquad \Delta\tau=\frac{1}{N}\,.
\end{equation}
recovers an inverted noise $x_0^{\text{inv}}$ that the model associates with $x_1^{\text{target}}$. The full procedure is in Algorithm~\ref{alg:paint}. Two implementation details matter:

\textbf{Constructing the target.}
We build $x_1^{\text{target}}$ by fixing its prefix positions to $A_{t-1}^{s:s+d-1}$
and filling the remaining positions with the tail $x_1^{\text{naive}}[d:]$ of a
standard forward pass from fresh noise, keeping $x_1^{\text{target}}$ on the data
manifold. Substituting zeros or random values for the free region moves it
off-manifold, destabilizing backward Euler integration, and corrupting
$x_0^{\text{inv}}$.


\textbf{Re-painting the free region.}
Inversion yields $x_0^{\text{inv}}[{:}d]$, encoding the prefix, and
$x_0^{\text{inv}}[d{:}]$, which we discard. Replacing the free region with fresh
noise $\varepsilon \sim \mathcal{N}(0, I)$ is tempting but problematic: token mixing
in $v_\pi$ spreads the mismatch between $\varepsilon$ and $x_0^{\text{inv}}[{:}d]$
across all positions during the final forward pass. Instead, we retain
$x_0^{\text{free}}[d{:}]$ from the naive pass, the same noise used to generate
$x_1^{\text{naive}}[d{:}]$ and hence the free region of $x_1^{\text{target}}$.
Since $x_0^{\text{free}}[d{:}]$ already encodes the same free output as
$x_0^{\text{inv}}[d{:}]$, substituting it introduces minimal disruption, following
the re-painting principle in~\citep{mao2023guided}.

The resulting chunk $A_{t} = \pi_\theta(x_0^\ast, o_t)$ satisfies the prefix constraint and produces a continuation that approximates the prefix-conditioned policy distribution:
\begin{equation}
    A_{t}^{0{:}d-1} \approx A_{t-1}^{s{:}s{+}d{-}1}  \quad \text{(prefix)};\ 
    A_{t}^{d:} \approx p_{\tau=1}\!\bigl(\cdot \mid A_{t-1}^{s{:}s{+}d-1}, o_t\bigr) \quad \text{(suffix)}\,.
\end{equation}
\section{Experiments}
\label{sec:experiments}

We design our experiments to answer two questions. First, \textit{``how does PAINT compare to existing inference methods in terms of both success rate and execution time?''}  Second, \textit{``how does the choice of the inversion method affect PAINT's performance?''}

\textbf{Metrics.} We evaluate two sets of metrics. The first measures task performance and efficiency: \textbf{s}uccess \textbf{r}ate (SR$ \uparrow $) and \textbf{a}verage \textbf{t}ime for successful \textbf{r}ollouts (ATR$ \downarrow $). The second assesses how well the generated chunk respects the prefix constraint via the \textbf{con}sistency score (CON$ \downarrow $). Formally, let $\mathcal{S}$ be a set of successful rollouts, $T_j$ the completion time of rollout $j$, and $d$ is the inference delay: 
\begin{align}
        \text{SR}
    =
    \frac{\#\text{successful\ trials}}
         {\#\text{total\ trials}}\,, \quad
    \text{ATR} =\frac{1}{|\mathcal{S}|}\sum_{j\in\mathcal{S}}T_j \,, \quad 
    \text{CON} =\frac{1}{d}\sum_{i=0}^{d-1}\left\|A_{t-1}[s+i]-A_t[i]\right\|_2
\end{align}

We provide more information about these metrics in the Appendix.

\vspace{-0.15in}
\subsection{Simulated Benchmark}
\vspace{-0.1in}
\textbf{Setup.}
We follow the exact setup of RTC~\cite{black2025real} on Kinetix~\citep{matthewskinetix}, using $12$ force-control environments and $4$-layer MLP-Mixer~\citep{tolstikhin2021mlp} flow policies with $H=8$. We report success rates (SR$ \uparrow $), and consistency scores (CON$ \downarrow $) with delays $d\in\{0,1,2,3,4\}$ and execution horizon $s\in\{d,\ldots,H-d\}$. 

We compare \texttt{PAINT} against five approaches, including (i) \textsc{naive async} (independent chunks), (ii) \textsc{temporal ensembling} (TE; overlap averaging)~\citep{zhao2023learning}, (iii) \textsc{B-spline refitting} (post-hoc smoothing)~\citep{yang2026abpolicy}, (iv) \textsc{BID (candidate filtering)}~\citep{liu2024bidirectional}, and (v) \textsc{RTC (gradient-based guidance)}~\citep{black2025real}. Additional details about these baselines are in~\Cref{apd:kinetix-baseline}.

\begin{figure}[t]
    \centering
    \includegraphics[width=\linewidth]{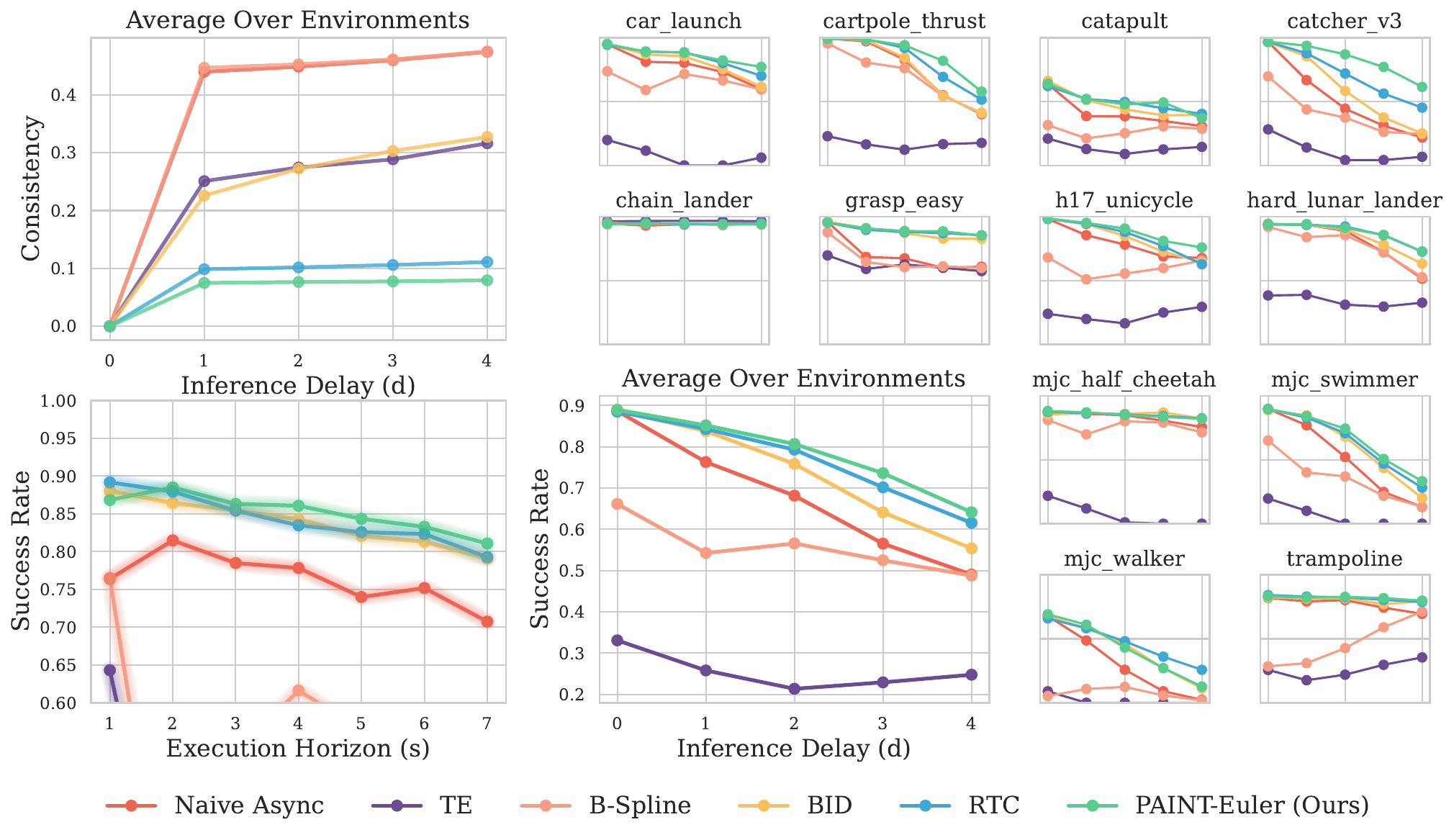}
    \caption{
    \textbf{Top left}: Inference delay ($d$) vs.\ prefix consistency (CON$\,\downarrow$)
    across all environments.
    \textbf{Bottom left}: Execution horizon vs.\ success rate (SR$\,\uparrow$) at $d{=}1$.
    \textbf{Right}: Inference delay ($d$) vs.\ success rate (SR$\,\uparrow$) across
    simulated $d \in \{0,1,2,3,4\}$.
    \texttt{PAINT}-Euler achieves the strongest overall performance across all delay values.
    Each data point aggregates $2048$ trials.}
    \label{fig:kinetix_exp}
\end{figure}

\begin{wrapfigure}[11]{R}{0.5\linewidth}
    \centering
    \includegraphics[width=\linewidth]{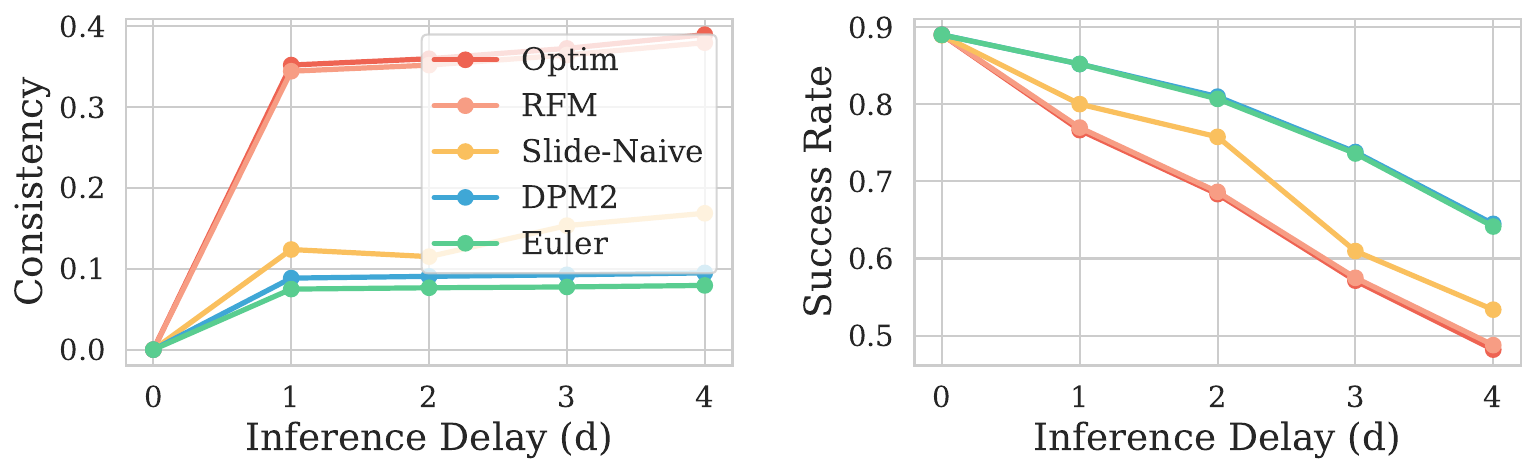}
    \caption{Average consistency scores and success rates over environments. Among various inversion methods, our chosen method (Euler) offers the best balance between quality and complexity.}
    \label{fig:ablation}
    \vspace{0.1in}
\end{wrapfigure}
\textbf{Results.}
~\Cref{fig:kinetix_exp} summarizes the simulated results. Under increasing delay, Naive Async degrades sharply, while TE and B-spline refitting provide limited robustness because averaging or smoothing generated actions does not explicitly condition on the executed prefix. B-spline refitting, in particular, leaves the prefix mismatch close to Naive Async, indicating that post-hoc smoothing alone is insufficient for chunk-boundary consistency. BID partially mitigates degradation, but remains below RTC and \texttt{PAINT} despite higher compute. \texttt{PAINT}-Euler achieves the strongest delay robustness without gradient computation and the lowest prefix mismatch, consistently improving over RTC. The execution-horizon sweep shows that \texttt{PAINT} and RTC benefit from shorter horizons because they can use more frequent feedback without introducing large chunk-boundary mismatches.

\textbf{Choices of Inversion Methods.}
We ablate the inversion step against four alternatives and a no-inversion baseline.
Backward Euler (Euler)~\citep{pan2023effective} runs $v_\theta$ in reverse for $N$
steps; DPM-Solver (DPM2)~\citep{hong2024exact} adds a midpoint correction for lower
error at twice the cost; single-step reverse flow matching (RFM)~\citep{li2026reverse}
exploits the linear interpolant at $t{=}1$ to recover noise in one call
($x_0^{\text{inv}} = x_1 - v_\theta(x_1^{\text{target}}, o, 1)$); and
optimization-based (Optim) inversion minimizes
$\|\pi_\theta(x_0^{\text{inv}}\mid o_t) - x_1^{\text{target}}\|^2$ via gradient
descent. Slide-Naive is a no-inversion baseline, which shifts the stored noise from the
previous inference forward by $d$ positions and resamples the free region.
\Cref{fig:ablation} shows that inversion quality correlates strongly with robustness
under asynchronous execution. Euler achieves the best balance of consistency,
stability, and efficiency -- on par with the costlier DPM2 -- while RFM, Optim, and
Slide-Naive exhibits a larger prefix mismatch and degrades more sharply with increasing
delay.

\subsection{Real-World Evaluation}

\begin{figure}[h]
\vspace{-0.05in}
    \centering
    \begin{subfigure}[b]{0.32\textwidth}
        \centering
        \includegraphics[width=\textwidth]{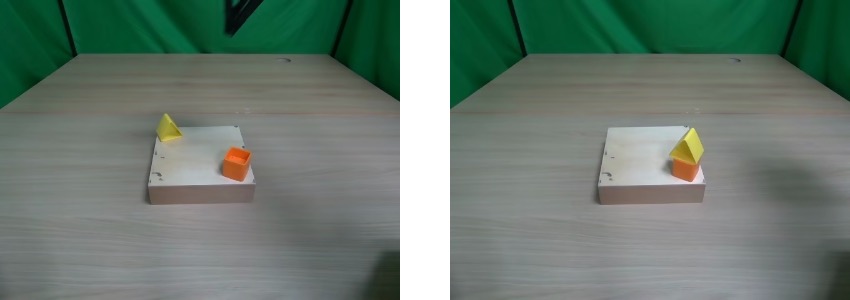}
        \caption{Block Stacking}
        \label{fig:real-a}
    \end{subfigure}
    \hfill
    \begin{subfigure}[b]{0.32\textwidth}
        \centering
        \includegraphics[width=\textwidth]{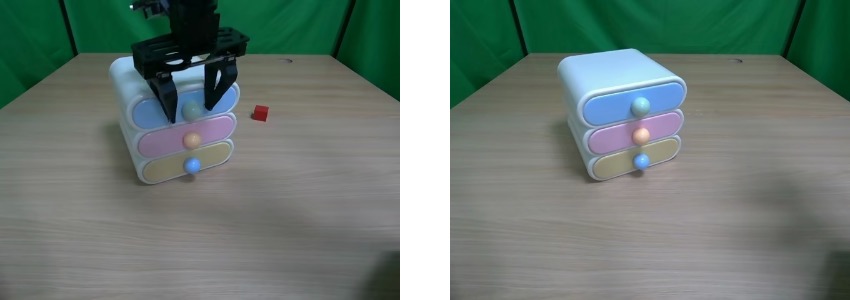}
        \caption{Toy in Drawer}
        \label{fig:real-b}
    \end{subfigure}
    \hfill
    \begin{subfigure}[b]{0.32\textwidth}
        \centering
        \includegraphics[width=\textwidth]{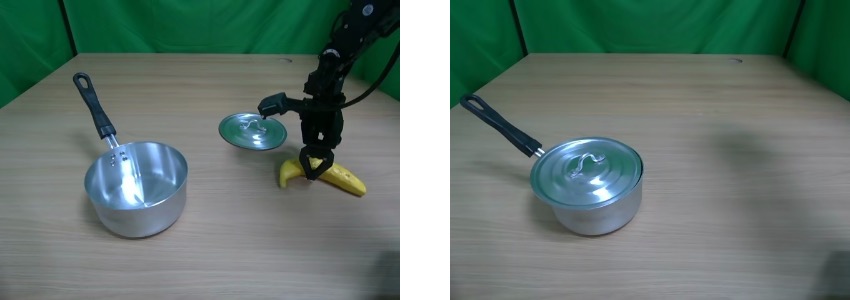}
        \caption{Banana in Pot}
        \label{fig:real-c}
    \end{subfigure}

    \begin{subfigure}[b]{0.32\textwidth}
        \centering
        \includegraphics[width=\textwidth]{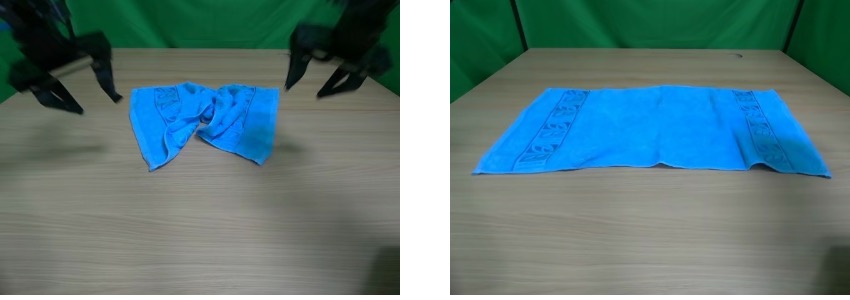}
        \caption{Towel Flinging}
        \label{fig:real-d}
    \end{subfigure}
    \hfill
    \begin{subfigure}[b]{0.32\textwidth}
        \centering
        \includegraphics[width=\textwidth]{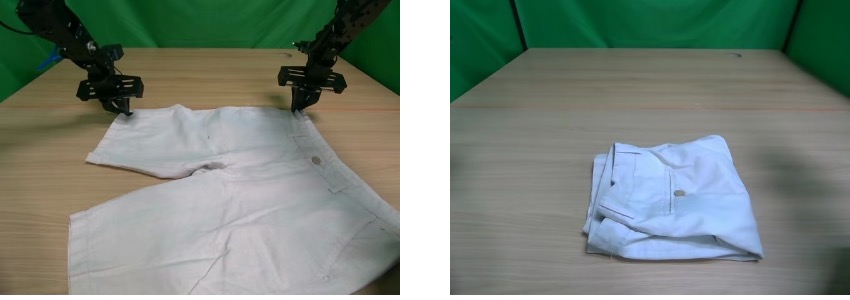}
        \caption{Shorts Folding}
        \label{fig:real-e}
    \end{subfigure}
    \hfill
    \begin{subfigure}[b]{0.32\textwidth}
        \centering
        \includegraphics[width=\textwidth]{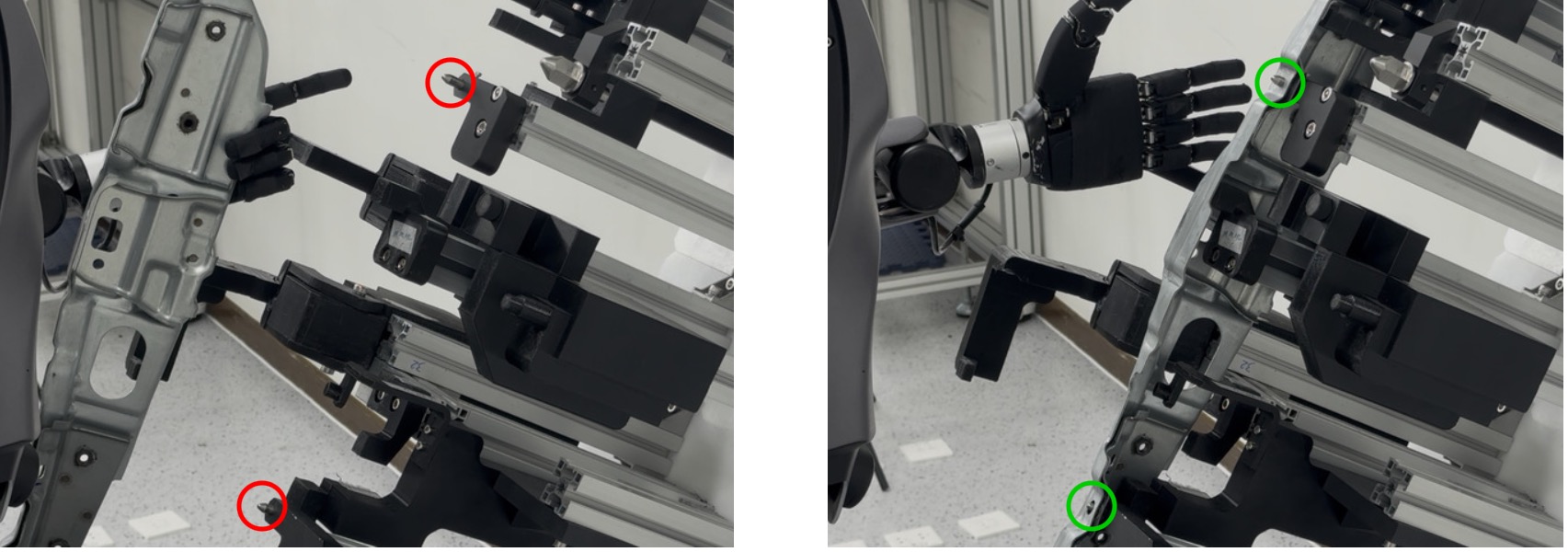}
        \caption{Part Placing}
        \label{fig:real-f}
    \end{subfigure}

    \caption{The environments for real-world evaluation. Each sub-figure's left image shows the initial state of environment; the right displays the goal state (more details are provided in the~\Cref{apd:real-task}).}
    \vspace{-0.1in}
    \label{fig:main}
\end{figure}

We next test whether \texttt{PAINT} transfers from controlled simulation to physical hardware across six real-world manipulation tasks spanning single-arm, bimanual (ALOHA~\citep{fu2024mobile}) and humanoid
settings (see~\Cref{fig:main}) and two VLA architectures (GR00T-N1.5~\citep{bjorck2025gr00t} and $\pi_0$~\citep{black2024pi_0}).
\label{sec:realworld}

\textbf{Setup.}
Our tasks: \textbf{(i) Block Stacking}, a precision pick-and-place task; \textbf{(ii) Toy in Drawer}
and \textbf{(iii) Banana in Pot}, single-arm multi-stage contact tasks; \textbf{(iv) Towel Flinging}
and \textbf{(v) Shorts Folding}, bimanual deformable-object tasks; and \textbf{(vi) Part Placing},
a humanoid precision-alignment task. All tasks except Towel Flinging use binary success scores; Towel Flinging is scored post-hoc based on towel flatness and spread. We use GR00T-N1.5~\citep{bjorck2025gr00t}
($H{=}16$, $N{=}4$) and $\pi_0$~\citep{black2024pi_0} ($H{=}50$, $N{=}10$) under remote
LAN inference, with $20$ trials per method--task pair, comparing against TE and RTC under
identical checkpoints and inference infrastructure. We focus on training-free inference-time baselines. Training-based approaches such as TT-RTC~\citep{black2025training} and A2C2~\citep{sendai2025leave} are orthogonal to \texttt{PAINT} as discussed in \Cref{apd:kinetix-ortho} and left
for future work. More details about setups are provided in~\Cref{apd:real-task}.
\begin{table}[t]
\centering
\scriptsize
\caption{Real-world evaluation over $20$ trials per method-task pair. $\pi_0$ results are omitted for Part Placing because the public checkpoint targets manipulator embodiments rather than the humanoid arm-hand platform. TE~\citep{zhao2023learning} is used as a synchronous inference and therefore does not have a CON score; we mark this entry as ``--'' in the table.}
\label{tab:real_world}
\begin{tabular}{l|ccc|ccc|ccc}
\toprule
 & \multicolumn{3}{c|}{\textbf{Block Stacking}}
 & \multicolumn{3}{c|}{\textbf{Toy in Drawer}}
 & \multicolumn{3}{c}{\textbf{Banana in Pot}} \\
\textbf{Method}
  & SR$\uparrow$ & ATR$\downarrow$ & CON$\downarrow$
  & SR$\uparrow$ & ATR$\downarrow$ & CON$\downarrow$
  & SR$\uparrow$ & ATR$\downarrow$ & CON$\downarrow$ \\
\midrule
GR00T-N1.5~\citep{bjorck2025gr00t} &&&&&&& \\
\quad TE~\citep{zhao2023learning} & 0.55 & 28.27 & -- & 0.60 & 23.19  & -- & 0.60 & 29.57 & -- \\
\quad RTC~\citep{black2025real}   & \textbf{0.75} & 16.04 & 0.030 & 0.75 & 17.85  & 0.025 & \textbf{0.70} & 30.15 & 0.031 \\
\rowcolor{blue!8}
\quad \textbf{PAINT (Ours)}       & \textbf{0.75} & \textbf{15.32} & \textbf{0.023} & \textbf{0.85} & \textbf{17.08} & \textbf{0.023} & \textbf{0.70} & \textbf{29.79} & \textbf{0.026} \\
\midrule
$\pi_0$~\citep{black2024pi_0} &&&&&&& \\
\quad TE~\citep{zhao2023learning} & 0.20 & 58.35 & -- & 0.15 & 53.27 & -- & 0.10 & 63.20 & -- \\
\quad RTC~\citep{black2025real}   & \textbf{0.50} & 15.30 & 0.036 & \textbf{0.65} & 36.35 & 0.032 & \textbf{0.55} & \textbf{37.41} & 0.028 \\
\rowcolor{blue!8}
\quad \textbf{PAINT (Ours)}       & \textbf{0.50} & \textbf{14.90} & \textbf{0.034} & \textbf{0.65} & \textbf{30.39} & \textbf{0.028} & \textbf{0.55} & 39.64 & \textbf{0.026} \\
\midrule\midrule
 & \multicolumn{3}{c|}{\textbf{Towel Flinging}}
 & \multicolumn{3}{c|}{\textbf{Shorts Folding}}
 & \multicolumn{3}{c}{\textbf{Part Placing}} \\
Method
  & SR$\uparrow$ & ATR$\downarrow$ & CON$\downarrow$
  & SR$\uparrow$ & ATR$\downarrow$ & CON$\downarrow$
  & SR$\uparrow$ & ATR$\downarrow$ & CON$\downarrow$ \\
\midrule
GR00T-N1.5~\citep{bjorck2025gr00t} &&&&&&& \\
\quad TE~\citep{zhao2023learning} & 0.51  & 17.13 & -- & 0.90 & 33.82 & -- & 0.50 & 68.51  & -- \\
\quad RTC~\citep{black2025real}   & 0.76 & \textbf{6.98} & 0.028 & 0.90 & \textbf{18.79} & 0.027 & \textbf{0.70}  & 18.28 & 0.030 \\
\rowcolor{blue!8}
\quad \textbf{PAINT (Ours)}       & \textbf{0.79} & 7.44 & \textbf{0.023} & \textbf{0.95} & 19.77 & \textbf{0.025} & \textbf{0.70} & \textbf{17.32} & \textbf{0.021} \\
\midrule
$\pi_0$~\citep{black2024pi_0} &&&&&&& \\
\quad TE~\citep{zhao2023learning} & 0.30 & 25.07 & -- & 0.40 & 52.32 & -- &&& \\
\quad RTC~\citep{black2025real}   & 0.54 & 17.04 & 0.039 & 0.65 & \textbf{29.64} & 0.028 & \multicolumn{3}{c}{N/A} \\
\quad\cellcolor{blue!8}\textbf{PAINT (Ours)}        &\cellcolor{blue!8}\textbf{0.80} &\cellcolor{blue!8}\textbf{15.16} &\cellcolor{blue!8}\textbf{0.027} &\cellcolor{blue!8}\textbf{0.70} &\cellcolor{blue!8}30.54 &\cellcolor{blue!8}\textbf{0.027} &&& \\
\bottomrule
\end{tabular}
\vspace{-0.1in}
\end{table}

\textbf{Results.}
\Cref{tab:real_world} summarizes the real-world evaluation across all tasks using GR00T-N1.5 and $\pi_0$. Across both architectures, \texttt{PAINT} generally matches or improves over RTC while requiring no backpropagation through the policy.
On GR00T-N1.5, \texttt{PAINT} improves the success rate on continuity-sensitive tasks such as Toy in Drawer, Towel Flinging, and Shorts Folding, and consistently improves CON when measured.
This suggests that selecting the initial noise before generation can produce smoother chunk transitions than steering the denoising trajectory after generation begins.
RTC enforces the prefix using gradient-based guidance~\citep{song2023pseudoinverse}, i.e., deployment-time backward operations, which can perturb the suffix after the prefix; on contact-rich or deformable-object tasks, such perturbations may produce small over-corrections or less stable contact behavior.
On $\pi_0$, \texttt{PAINT} remains competitive on single-arm tasks and improves performance on bimanual deformable-object manipulation, indicating transferability across flow-based VLA architectures.

Compared with TE, \texttt{PAINT} substantially reduces rollout time while preserving prefix consistency.
TE averages overlapping chunks, which can smooth motion but may also blend conflicting commands. This issue worsens with larger $H$ in retry cases, where older actions may continue forward while the new chunk corrects backward, pulling the final command toward the middle.
Finally, \texttt{PAINT} is faster than RTC on GR00T-N1.5 ($86 {\pm}2$ vs.\ $113{\pm}3$~ms) but slower on $\pi_0$ ($311{\pm}7$ vs.\ $213{\pm}4$~ms), reflecting its core trade-off: replacing gradient-based correction with additional forward-only evaluations.
\vspace{-0.1in}
\section{Conclusion}
\label{sec:conclusion}

\vspace{-0.1in}
\texttt{\texttt{PAINT}} shows that asynchronous action-chunk execution can be addressed before generation
begins, by selecting an initial noise that anchors the next chunk to the actions already
executed. This avoids retraining and deployment-time gradient correction while preserving
the base policy dynamics unchanged. Our experiments suggest that noise-space adaptation is a
practical mechanism for improving chunk-boundary consistency in flow-based robot policies.
\texttt{PAINT} is relevant when asynchronous execution requires prefix consistency and deployment-time gradient computation is undesirable or unsupported. Its additional forward evaluations introduce a computational trade-off, and its latency relative to RTC depends on the policy architecture and solver configuration.
Because \texttt{PAINT} uses only forward model evaluations, it is also more naturally compatible with graph-compiled deployment pipelines such as TensorRT than methods that require deployment-time vector-Jacobian products. More broadly, our results highlight the initial
noise as a useful control interface~\citep{patil2026you, wagenmakersteering,
pan2025adonoisingdispellingmyths, ye2025ra} rather than a passive random input, enabling real-time adaptation of generative robot policies without modifying the learned generative process.


\section{Limitations}
\vspace{-0.1in}
\label{sec:limitations}
\texttt{PAINT} relies on two practical assumptions, each suggesting a direction for future work.
First, it benefits from approximate locality between noise and action positions:
changing the prefix region of the initial noise should mostly affect the prefix of the
generated chunk. This locality is encouraged by optimal-transport flow matching, but
may weaken for architectures with strong cross-position mixing or highly multimodal
action distributions. Future work could quantify locality in deployed VLA backbones via
noise-perturbation probes and, where it is weak, augment \texttt{PAINT} with a token-attention
mask or a brief locality-preserving fine-tuning stage. Second, \texttt{PAINT} uses backward
Euler inversion, which works well when the sampling trajectory is close to straight.
This is reasonable for OT flow matching with a linear interpolant, but may be less
accurate for variance-preserving diffusion models, whose probability paths are more
curved. Extending \texttt{PAINT} to such models would likely require a dedicated inversion
procedure (e.g., DDIM inversion or higher-order solvers such as DPM-Solver), possibly
with a learned correction term for residual discretization error. Finally, our real-world evaluation uses a single natural inference-delay setting ($d \approx 3$); evaluating a broader range of delays on physical hardware would further characterize robustness.

\section*{Acknowledgements}
Duy M. H. Nguyen acknowledges the support of the Federal Ministry of Research, Technology, and Space (BMFTR) under grant number 16IW26002 (AMENABLE) and the Federal Ministry of Education and Research (BMBF) under grant number 16IW24006 (NoIDLEChatGPT).




\bibliography{example}  
\clearpage
\section*{Appendix Contents}
\startcontents[appendix]
\printcontents[appendix]{l}{1}{\setcounter{tocdepth}{2}}
\clearpage

\appendix
\renewcommand\thesection{\Alph{section}}
\crefalias{section}{appendix}
\crefalias{subsection}{appendix}

\section{Kinetix Benchmark}
\begin{figure}[h]
    \centering
    \includegraphics[width=\linewidth]{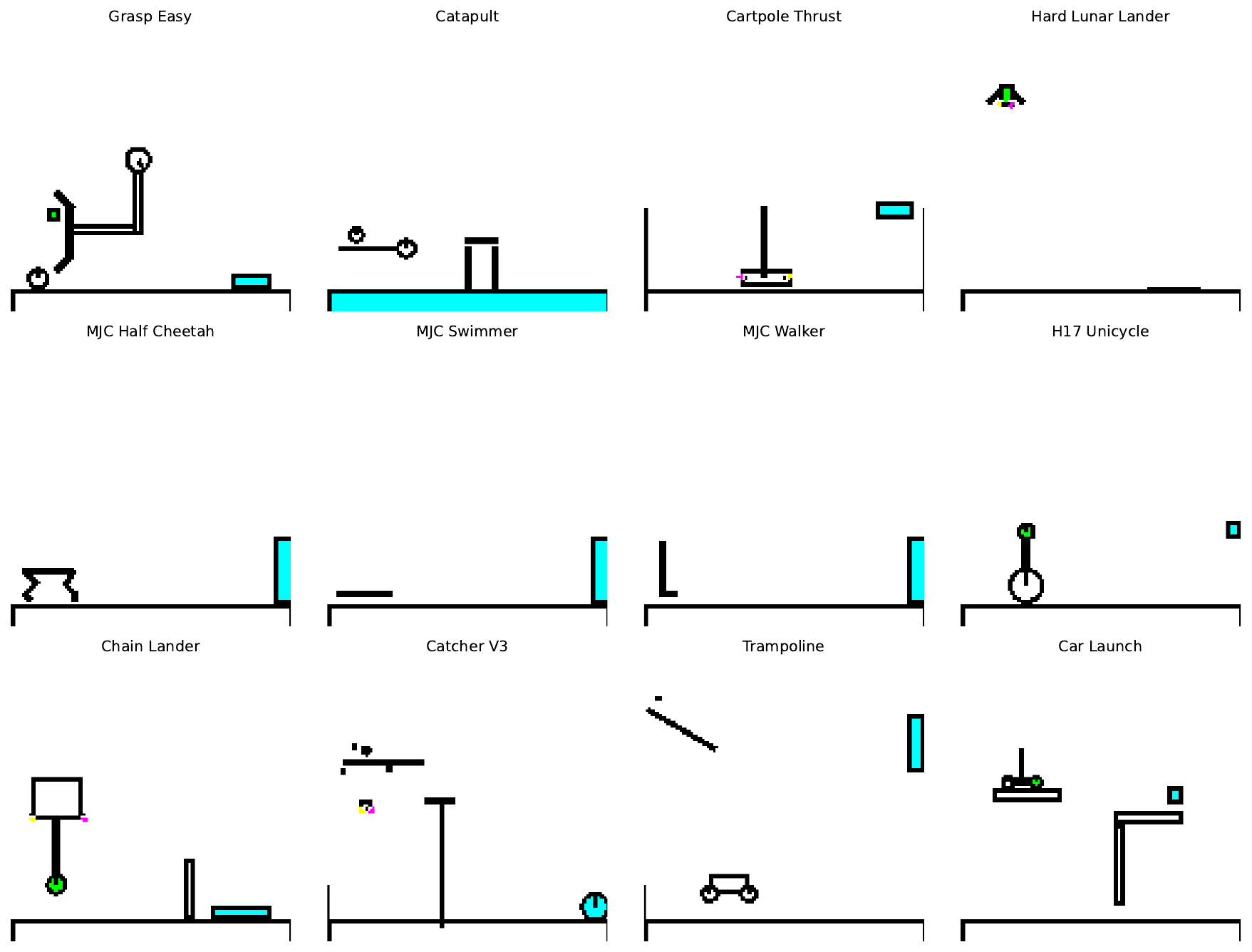}
    \caption{The visualizer of $12$ environments in Kinetix~\citep{matthewskinetix}.}
    \label{fig:kinetix_v_apd}
\end{figure}

Kinetix~\citep{matthewskinetix} is an open-ended, two-dimensional rigid-body
physics benchmark designed for evaluating generalist agents across a broad
range of control problems.
Each Kinetix environment is procedurally defined by a configuration of
polygons, circles, joints, and thrusters governed by a 2D physics simulator,
yielding a unified action and observation space across drastically different
tasks.
Each environment also defines a binary success criterion specified through
colour-coded entities (e.g.\ green target circles, red obstacles to avoid)
and goal joints that must be activated, so success is determined
automatically by the simulator.
This makes Kinetix particularly well-suited for evaluating asynchronous
inference methods: the same policy architecture and learning pipeline must
operate over locomotion, manipulation, control, projectile dynamics
without task-specific modifications, and a single success metric is
directly comparable across the entire benchmark. See~\Cref{fig:kinetix_v_apd}.

\subsection{Baseline Details}
\label{apd:kinetix-baseline}

We compare \texttt{PAINT} against representative baselines for sync/asynchronous
action-chunk execution.
Unless otherwise stated, all methods use the same pretrained policy
checkpoint and differ only in how action chunks are executed, filtered,
or modified at chunk boundaries.

\textbf{Naive Async.}
The robot executes the current chunk while the next chunk is generated
in parallel, and then switches directly to the new chunk as soon as it is
fully available.
Each query uses independently sampled initial noise, and no constraint
is imposed between the executed actions and the prefix of the new chunk.
This method is cheap and requires no retraining or gradients, but can
create discontinuities at chunk boundaries.

\textbf{Temporal Ensembling (TE)}~\citep{zhao2023learning} averages
overlapping action predictions across chunks, following the aggregation
strategy of ACT.
In recovery scenarios, older chunks may perpetuate the previous motion
while newer ones attempt correction; averaging these conflicting
predictions can bias the action toward an intermediate trajectory,
slowing task completion.

\textbf{B-spline Refitting.}
We use only the post-hoc trajectory-refitting component inspired by
ABPolicy~\citep{yang2026abpolicy}, without retraining the base policy
or changing its action representation.
The generated chunk is refit with a smooth B-spline to reduce abrupt
transitions, testing whether smoothness alone is sufficient for
asynchronous execution.
Since the refitting does not condition generation on the executed
prefix, it can reduce visual jerkiness while still creating a prefix
mismatch.

\textbf{Bidirectional Decoding (BID)}~\citep{liu2024bidirectional}
samples multiple candidate chunks and selects one according to a
consistency or scoring criterion.
It requires no gradients or retraining, but increases inference cost
because multiple chunks are generated per query.
Unlike our method, BID addresses asynchronous mismatch by filtering
generated outputs rather than modifying the denoising process or
selecting the initial noise directly.

\textbf{Real-Time Chunking (RTC)}~\citep{black2025real} steers the
denoising trajectory during generation to satisfy the prefix constraint.
At each denoising step, it uses vector-Jacobian products (VJPs) to
move the predicted final chunk toward the executed prefix.
RTC directly targets chunk-boundary consistency, but requires
deployment-time backward operations through the policy, which may be
expensive or unsupported in graph-compiled runtimes.

\textbf{Training-time RTC (TT-RTC)}~\citep{black2025training} modifies
the policy during training so that the learned model becomes more
tolerant to inference delay.
It requires fine-tuning but does not require VJPs at deployment.
We include TT-RTC to distinguish training-time adaptation from \texttt{PAINT}'s
inference-time noise selection, and additionally evaluate
\textbf{TT-RTC + \texttt{PAINT}} to test whether prefix-anchored noise selection
can further improve a delay-aware policy.

\textbf{A2C2}~\citep{sendai2025leave} is an inference-time
action-correction method that uses the recent execution context to improve
chunk-boundary consistency without retraining the base policy.
It operates in action space after generation, while \texttt{PAINT} acts earlier
by selecting the initial noise.
\textbf{A2C2 + \texttt{PAINT}} uses A2C2's correction mechanism to chunks generated from \texttt{PAINT}'s prefix-anchored noise. The goal is to test whether noise-space anchoring and action-space correction
provide complementary benefits.

\subsection{PAINT vs. Training-Time Methods}
\label{apd:kinetix-ortho}

\begin{figure}[h]
    \centering
    \includegraphics[width=\linewidth]{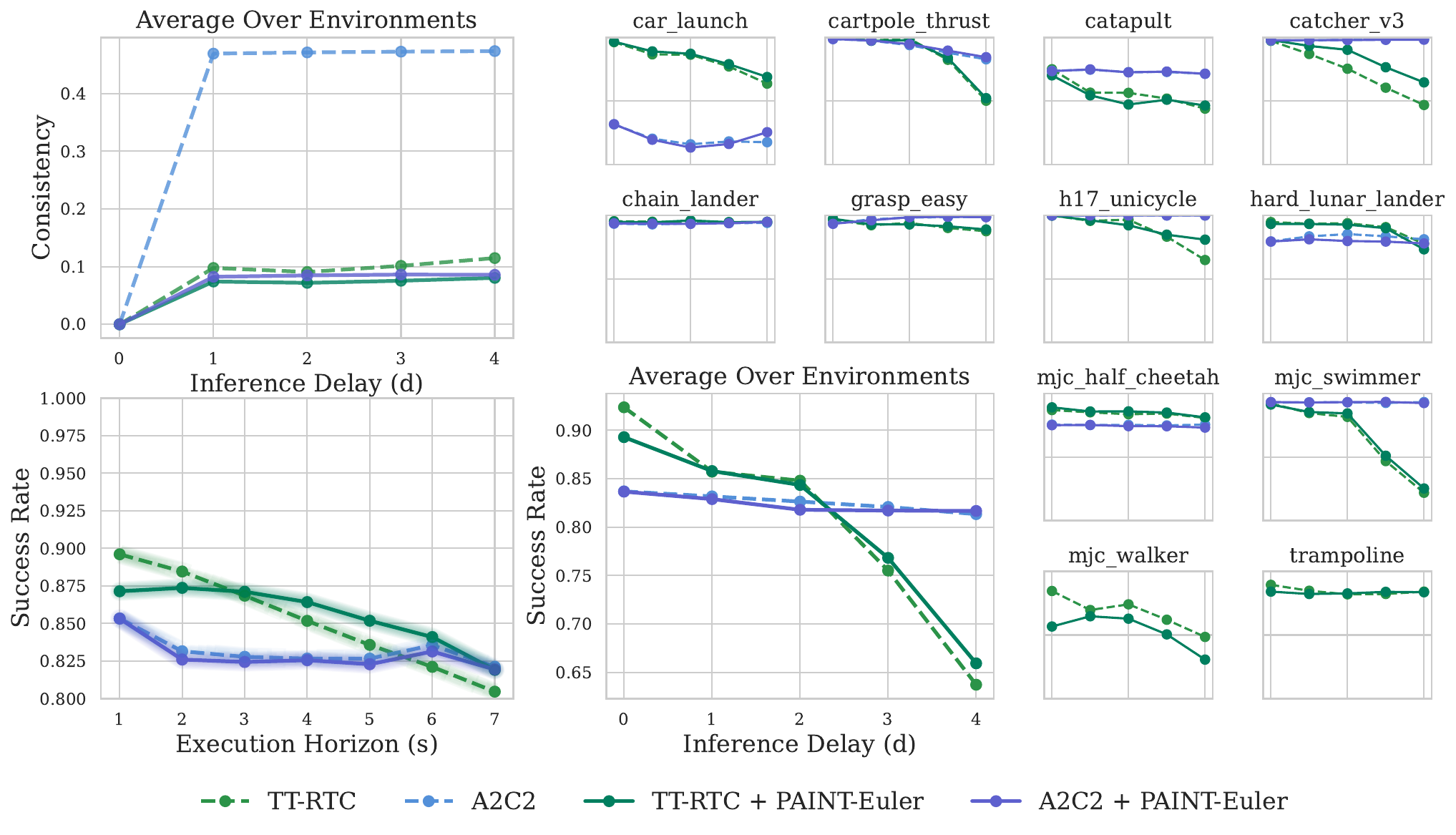}
    \caption{\texttt{PAINT} v.s. training-time delay-aware
    methods on the Kinetix benchmark. We evaluate \texttt{PAINT} on top of
    A2C2~\citep{sendai2025leave} and training-time
    RTC (TT-RTC)~\citep{black2025training} , measuring both consistency
    score and task success rate under varying execution
    horizons and inference delays.}
    \label{fig:kinetix_apd}
\end{figure}

A2C2~\citep{sendai2025leave} and TT-RTC~\citep{black2025training}
improve asynchronous execution by modifying the policy during training,
whereas \texttt{PAINT} operates purely at inference time by selecting a
prefix-anchored initial noise.
We therefore evaluate \texttt{PAINT} on top of both trained policies to test
whether inference-time noise selection is complementary to training-time
delay adaptation.
As shown in \Cref{fig:kinetix_apd}, \texttt{PAINT} substantially reduces prefix
mismatch for both A2C2 and TT-RTC across all nonzero delays.
Adding \texttt{PAINT} to TT-RTC decreases prefix consistency error (CON) from
$0.11$ to $0.08$ at $d=4$ while preserving or improving task success,
and adding \texttt{PAINT} to A2C2 similarly reduces CON by approximately $80\%$.
In terms of task success, adding \texttt{PAINT} largely preserves A2C2's
performance across execution horizons and delays, while TT-RTC+\texttt{PAINT}
achieves comparable success at small delays and greater robustness at
larger delays.
These results indicate that \texttt{PAINT} is orthogonal to training-time
adaptation: it can be applied on top of existing delay-aware policies
to improve prefix consistency and, in some settings, high-delay
robustness, without additional retraining.

\section{Real-Robot Environment Details}
\label{apd:real-task}

\begin{figure}
  \begin{subfigure}{\linewidth}
    \includegraphics[width=\linewidth]{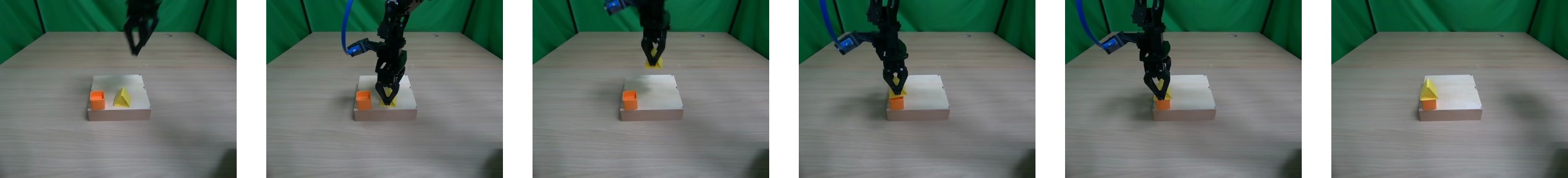}
    \caption{Block Stacking}
  \end{subfigure}
  \begin{subfigure}{\linewidth}
    \includegraphics[width=\linewidth]{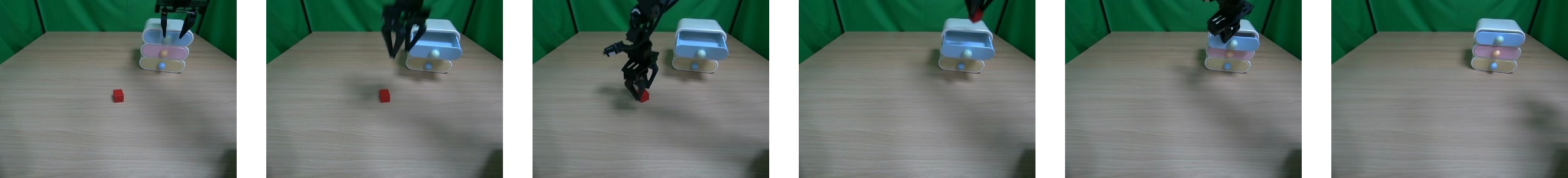}
    \caption{Toy in Drawer}
  \end{subfigure}
  \begin{subfigure}{\linewidth}
    \includegraphics[width=\linewidth]{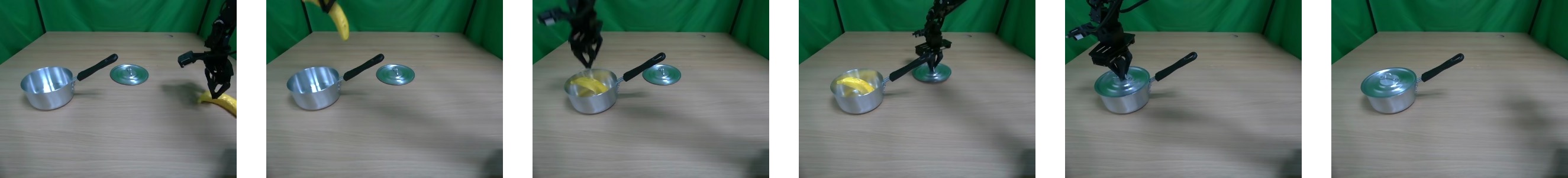}
    \caption{Banana in Pot}
  \end{subfigure}
  \begin{subfigure}{\linewidth}
    \includegraphics[width=\linewidth]{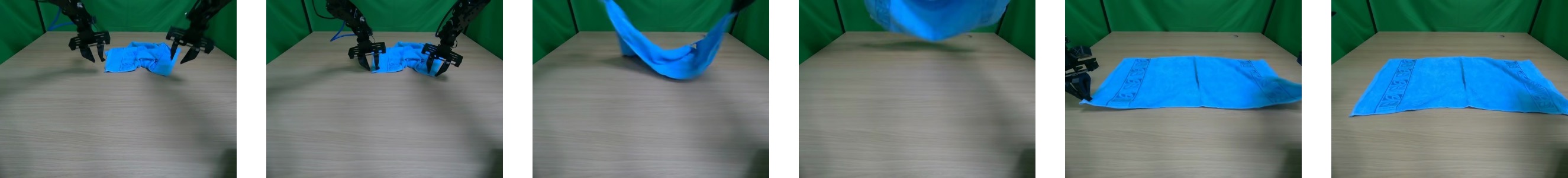}
    \caption{Towel Flinging}
  \end{subfigure}
  \begin{subfigure}{\linewidth}
    \includegraphics[width=\linewidth]{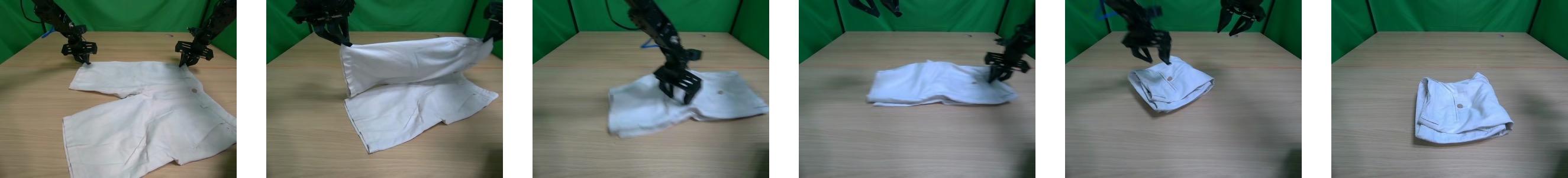}
    \caption{Shorts Folding}
  \end{subfigure}
  \begin{subfigure}{\linewidth}
    \includegraphics[width=\linewidth]{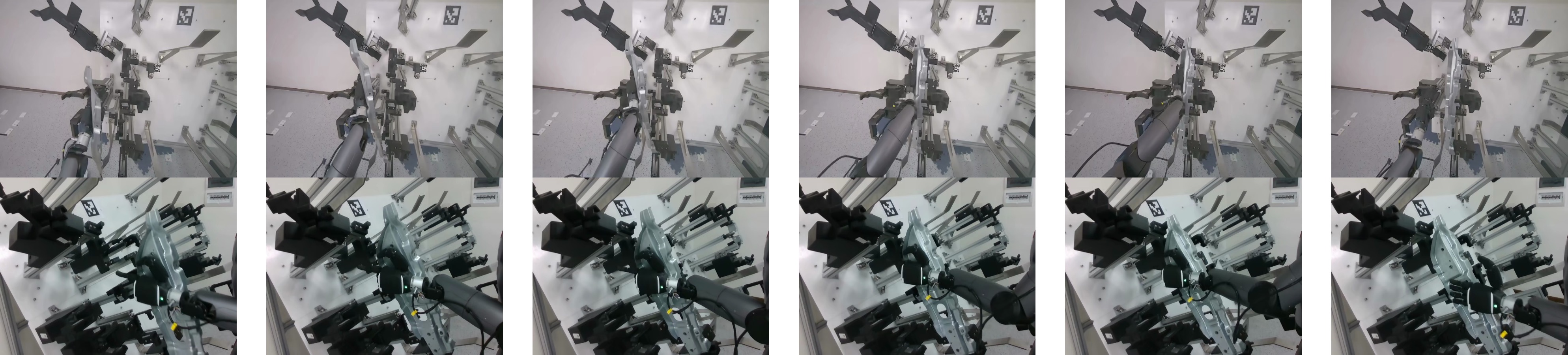}
    \caption{Part Placing}
  \end{subfigure}
  \caption{Examples of completed sequences in real-world tasks.}
  \label{fig:real_world_tasks}
\end{figure}

We evaluate \texttt{PAINT} on six real-world manipulation tasks spanning
single-arm manipulation, bimanual deformable-object manipulation, and
humanoid part placement (see~\Cref{fig:real_world_tasks}).
All methods are evaluated using the same policy checkpoint, robot
hardware, camera observations, action horizon, and inference
infrastructure; the only difference is the inference-time execution
strategy.

\begin{table}[h]
\centering
\caption{Real-world task descriptions.}
\label{tab:real_robot_details}
\begin{tabular}{llp{3.0in}}
\toprule
\textbf{Task} & \textbf{Embodiment} & \textbf{Task Description} \\
\midrule
Block Stacking & Single-arm &
A precision pick-and-place task requiring the robot to grasp a
triangular block and stack it stably on top of a square block. \\

Toy in Drawer & Single-arm &
A multi-stage contact task involving drawer opening, cube placement
inside the drawer, and drawer closing. \\

Banana in Pot & Single-arm &
A sequential object-manipulation task requiring the robot to place a
banana into a pot and then close the pot with a lid. \\

Towel Flinging & Bimanual &
A dynamic deformable-object manipulation task requiring coordinated
bimanual grasping, lifting, and flinging to spread a towel flat on
the table. Performance is reported as a post-hoc score from $0$ to $10$;
the score measures the extent to which the towel covers the table,
with flatter configurations receiving higher scores. \\

Shorts Folding & Bimanual &
A bimanual deformable-object task requiring coordinated folding of
the shorts into a compact final configuration. \\

Part Placing & Humanoid arm-hand &
A precision-alignment task requiring the humanoid robot to place a
part onto a two-pin jig with accurate pose alignment. \\
\bottomrule
\end{tabular}
\end{table}

\textbf{Hardware Setup.}
For ALOHA~\citep{fu2024mobile} experiments, the robot consists of two
ViperX $6$-DoF arms, each equipped with a parallel-jaw gripper, for
a total of $14$ controllable degrees of freedom ($12$ arm + $2$ gripper).
We use a single-arm configuration ($7$ DoFs: $6$ arm + $1$ gripper)
for rigid-object manipulation tasks and the full bimanual configuration
($14$ DoFs) for deformable-object manipulation.
For humanoid experiments, we use a robot with a single arm of $7$ DoFs and a $6$-DoF
dexterous hand.
The policy receives $640 \times 480$ RGB observations from $3$ cameras,
proprioceptive joint positions, and language/task conditioning when
required by the base VLA.

\textbf{Control and Inference.}
The controller runs at $20$\,Hz on a workstation
equipped with an NVIDIA RTX~4070 ($16$\,GB), AMD Ryzen~9~7900X,
and 64~GB RAM, running Ubuntu~24.04.
Each policy call predicts an action chunk of horizon $H$, executed
asynchronously while the next chunk is generated, yielding a natural
inference delay of $d = \lfloor \delta / \Delta t \rfloor$ where
$\delta$ is the wall-clock inference time including network transfer.

\textbf{Evaluation Procedure.}
Each method-task pair is evaluated over $20$ trials with random
initial-state distributions.
We report success rate (SR), average time for successful rollouts (ATR),
and prefix consistency error (CON) when applicable.
A trial is marked as failed if the robot violates the task success
criterion, loses contact with the target object in an unrecoverable
way, enters an unsafe configuration, or reaches the maximum episode
length. We provide the task description in~\Cref{tab:real_robot_details}.

\section{Inversion Methods}
\label{apd:inversion_comparison}

We compare several ways to recover a prefix-anchored initial noise.
All variants construct a target endpoint $x^{\text{target}}_1$ by
combining the executed prefix with a suffix generated from a naive
forward pass.
They differ only in how they recover the corresponding initial noise. \emph{The number of model calls in Table~\ref{tab:inversion_methods}
refers to the inversion step alone}; total \texttt{PAINT} inference cost
(including the naive forward pass and final forward pass) is reported
in Table~\ref{tab:inference_methods}.

\begin{table}[h]
\centering
\caption{Comparison of inversion methods used in \texttt{PAINT} ablations.
$N$ denotes the number of flow-matching solver steps.}
\label{tab:inversion_methods}
\resizebox{\linewidth}{!}{
\begin{tabular}{llccl}
\toprule
Method & Description & Model Calls (inversion only) & Gradients? & Expected Behavior \\
\midrule
Backward Euler  & Reverse ODE integration              & $N$           & No  & Stable, simple, good cost--quality trade-off \\
DPM2 / midpoint & Higher-order reverse integration     & $\approx 2N$  & No  & Lower discretization error, higher cost \\
Single-step RFM & Linear-interpolant inverse at $\tau=1$ & $1$         & No  & Fast but less accurate for curved trajectories \\
Optimization    & Minimize endpoint reconstruction error & $\gg N$ + grad. & Yes & Accurate but slow and deployment-unfriendly \\
Slide-Naive     & Shift stored noise and resample suffix & $\approx 0$ (amortized) & No & Cheap but accumulates stale-noise error \\
\bottomrule
\end{tabular}
}
\end{table}

\textbf{Backward Euler.}
The default \texttt{PAINT} implementation starts from $x^{\text{target}}_1$
and integrates backward:
\begin{equation}
    x_{\tau-\Delta\tau} = x_{\tau} - \Delta\tau \, v_\theta(x_\tau, o, \tau)\,.
\end{equation}
This requires $N$ model evaluations and no gradients.

\textbf{DPM2 / Midpoint Inversion.}
A higher-order solver reduces local truncation error by evaluating
the vector field at an intermediate point.
This can improve inversion accuracy when the vector field is curved,
but approximately doubles the number of model evaluations.

\textbf{Single-step RFM Inversion.}
For flow matching with a linear interpolant, a single-step estimate
can be obtained near $\tau=1$:
\begin{equation}
    x_0^\ast \approx x_1^{\text{target}} - v_\theta(x_1^{\text{target}}, o, 1)\,.
\end{equation}
This is computationally attractive but can be inaccurate when the
learned trajectory is nonlinear or when the target prefix lies in a
high-curvature region of the action distribution.

\textbf{Optimization-based Inversion} optimizes $x_0$ directly:
\begin{equation}
    \min_{x_0}
    \left\|
    \pi_\theta(x_0, o) - x_1^{\text{target}}
    \right\|_2^2\,.
\end{equation}
This can produce accurate inversions but requires repeated forward
passes and gradients through the policy, making it less suitable for
latency-critical deployment.

\textbf{Slide-Naive} shifts the stored initial noise forward by $d$
positions and samples fresh noise for the suffix.
This is inexpensive, but the shifted noise may become stale as the
observation and action distributions evolve over time.

\section{Invertibility, Discretization Error, and Stability}
\label{apd:invertibility_stability}

\texttt{PAINT} relies on approximately inverting the learned flow dynamics from
a target action chunk to an initial noise.
For continuous-time flow matching, the ODE
\begin{equation}
    \frac{dx_\tau}{d\tau} = v_\theta(x_\tau, o, \tau)\,,
    \qquad \tau \in [0,1]\,,
\end{equation}
admits a unique solution under standard regularity conditions, such
as Lipschitz continuity of $v_\theta$ in $x_\tau$ and continuity in
$\tau$~\citep{lipman2024flow}.
Under these conditions, the continuous flow map is invertible.
However, practical policies are evaluated with finite-step numerical
solvers, so the inverse recovered by \texttt{PAINT} is only approximate.

\textbf{Discretization Error.}
\texttt{PAINT}-Euler uses backward Euler integration:
\begin{equation}
    x_{\tau-\Delta \tau} = x_\tau - \Delta \tau \, v_\theta(x_\tau, o, \tau)\,.
\end{equation}
This introduces a discretization error that depends on the step size
$\Delta \tau = 1/N$, the local curvature of the learned vector field,
and the numerical stability of the reverse integration.
When $N$ is small, the step size is large, and the recovered noise may
deviate from the true inverse trajectory, increasing the prefix mismatch
after the final forward pass.
A larger $N$ improves the inversion accuracy but increases the inference cost.

\textbf{Stability Considerations.}
Reverse integration is stable when the solver step size is sufficiently
small relative to the local Lipschitz constant of the vector field.
If $\Delta \tau$ is too large, the Euler inversion can overshoot in regions
where $v_\theta$ changes rapidly, especially for highly multimodal
action distributions or architectures with strong cross-token mixing.
In practice, we find that using the same number of reverse steps as
the forward inference solver provides a good trade-off between
stability and cost.

\textbf{Approximate Inversion in Practice.}
Our method does not require the discretized policy to be exactly
invertible.
It only requires that backward integration recovers an initial noise
whose forward rollout produces a prefix sufficiently close to the
executed prefix.
\texttt{PAINT} should therefore be interpreted as an approximate inverse
procedure for prefix anchoring, not as an exact inverse of the learned
policy.
Notably, all backward integration steps are gradient-free forward
evaluations of $v_\theta$, making \texttt{PAINT} compatible with compiled
inference runtimes such as TensorRT.

\section{Inference Method Comparison}
\label{apd:inference_method_comparison}

Table~\ref{tab:inference_methods} compares \texttt{PAINT} with representative
inference-time and training-time approaches for asynchronous
action-chunk execution.
Model-call counts in this table refer to the \emph{full inference
pipeline}, including any naive forward pass, inversion, and final
forward pass.

\begin{table}[h]
\centering
\caption{Comparison of asynchronous inference methods. $N$ denotes the
number of denoising or flow-matching steps, $B$ denotes the number of
candidate chunks in rejection-sampling methods, and VJP denotes
vector-Jacobian product.}
\label{tab:inference_methods}
\resizebox{\linewidth}{!}{
\begin{tabular}{lcccl}
\toprule
Method & Retraining? & Gradients? & Approx. Model Evaluations & Notes \\
\midrule
Naive Async       & No  & No  & $N$ & Independent noise per chunk; no prefix enforcement \\
Temporal Ensembling & No & No  & $N$ per query & Dense re-querying with overlapping chunks \\
BID               & No  & No  & $B N$ & Rejection sampling over candidate chunks \\
RTC               & No  & Yes & $N$ (forward + VJP) & Steers velocity field at each denoising step \\
TT-RTC            & Yes & No  & $N$ & Trains policy to tolerate delay \\
\texttt{PAINT}-RFM         & No  & No  & $N + 1$ & Single-step inverse, then final forward pass \\
\texttt{PAINT}-Euler       & No  & No  & $3N$ & Naive forward, backward Euler inversion, final forward \\
\texttt{PAINT}-DPM2        & No  & No  & $\approx 4N$ & Higher-order inversion, higher cost \\
\texttt{PAINT}-Optim       & No  & Yes & $\gg N$ + grad. & Direct latent optimization; slow but diagnostic \\
\texttt{PAINT}-Slide-Naive & No  & No  & $\approx N$ & Reuses shifted noise after initial inversion \\
\bottomrule
\end{tabular}
}
\end{table}

\textbf{Deployment Implications.} RTC and optimization-based inversion require gradients or VJPs through
the policy during deployment.
This may be difficult to support in graph-compiled inference runtimes
or systems optimized only for forward execution.
In contrast, \texttt{PAINT}-Euler requires only forward evaluations of the
velocity network, making it easier to integrate with deployment
runtimes such as TensorRT or other graph-mode accelerators.

\textbf{Compute Trade-off.}
\texttt{PAINT}-Euler is more expensive than naive asynchronous inference in
raw model calls, but avoids backpropagation and improves prefix
consistency.
\texttt{PAINT}-Slide provides a lower-cost variant by reusing and shifting the
recovered noise across consecutive chunks, at the cost of possible
noise staleness.
This creates a practical trade-off between consistency, latency, and
deployment simplicity.

\section{Repainting Design}
\label{apd:repainting_ablation}

The term \emph{repainting} refers to replacing the prefix portion of
the initial noise, not the suffix.
The suffix noise $x_0^{\text{free}}[d:]$ is sampled from the prior
$\mathcal{N}(0, I)$ at the very beginning and is never changed.
What motivates the procedure is that a naive forward pass from
$x_0^{\text{free}}$ produces a chunk $x_1^{\text{naive}}$ whose prefix
does not satisfy the prefix constraint.
Rather than discarding the entire noise and starting over, we keep
the suffix noise intact --- it already encodes a valid continuation
under the current observation --- and \emph{repaint only the prefix
noise} $x_0^{\text{free}}[:d]$ with the inverted $x_0^*[:d]$ that
anchors the output to the executed prefix.
The final repainted noise is:
\begin{equation}
x_0^{\text{repaint}}
=
\bigl[\;
\underbrace{x_0^*[:d]}_{\text{repainted}},
\;\;
\underbrace{x_0^{\text{free}}[d:]}_{\text{original}}
\;\bigr].
\end{equation}

\textbf{Why Discarding $x_0^*[d:]$?}
The inverted suffix $x_0^*[d:]$ was recovered from a target
$x_1^{\text{target}}$ whose prefix was manually replaced.
Although it produces a valid reverse trajectory for this constructed
target, it aligns too closely with the artificial target endpoint and
reduces suffix diversity.
Repainting the prefix with keeping $x_0^{\text{free}}[d:]$ instead preserves the
stochastic continuation of the original policy while anchoring only
the prefix.

\textbf{Why Keeping $x_0^{\text{free}}[d:]$ Rather Than Fresh Noise?}
The suffix noise was sampled from the correct prior and used to
generate $x_1^{\text{naive}}[d:]$, which is also the free region of
$x_1^{\text{target}}$ that we inverted.
The model already associates $x_0^{\text{free}}[d:]$ with a valid
on-manifold continuation under observation $o$.
Replacing it with fresh noise $\varepsilon \sim \mathcal{N}(0, I)$
introduces a discontinuity at position $d$: the token mixing layers
of $v_\theta$ propagate this mismatch between $x_0^*[:d]$ and
$\varepsilon[d:]$ across all positions during the final forward pass,
degrading prefix quality.
Keeping the original suffix avoids this coupling error.

\section{Evaluation Metrics}
\label{apd:metrics}

\textbf{Success Rate (SR)} is the fraction of trials that satisfy the task-specific
success criterion:
\begin{equation}
    \text{SR} =
    \frac{\#\text{successful trials}}{\#\text{total trials}}\,.
\end{equation}

\textbf{Average Time for Successful Rollouts (ATR)} is the mean completion time over successful trials:
\begin{equation}
    \text{ATR} =
    \frac{1}{|\mathcal{S}|} \sum_{i \in \mathcal{S}} T_i\,,
\end{equation}
where $\mathcal{S}$ is the set of successful trials and $T_i$ is the
completion time of trial $i$.

\textbf{Prefix Consistency Error (CON)} measures how closely the newly generated chunk matches the actions
already executed during inference:
\begin{equation}
    \text{CON} =
    \frac{1}{d} \sum_{i=0}^{d-1}
    \left\| A_t[i] - A_{t-1}[s+i] \right\|_2\,.
\end{equation}

For real-world experiments, actions are reported in joint-angle space
(radians) without per-dimension weighting; rotational and translational
dimensions are treated equally.
Lower CON indicates better satisfaction of the prefix constraint.

\end{document}